\documentclass[letterpaper, 10 pt, conference]{ieeeconf}  % Comment this line out if you need a4paper
\usepackage[
           CJKbookmarks=true,
           bookmarksnumbered=true,
           bookmarksopen=true,
           colorlinks=false,
           pdfborder=0 0 1,
           citecolor=blue
           ]{hyperref}
\usepackage{graphics} % for pdf, bitmapped graphics files
\usepackage[T1]{fontenc}
\usepackage{mathptmx} % assumes new font selection scheme installed
\usepackage{times} % assumes new font selection scheme installed
\usepackage{amsmath} % assumes amsmath package installed
\usepackage{amssymb}  % assumes amsmath package installed
\usepackage{subfig}
\usepackage{dblfloatfix}
\usepackage{multirow}
\usepackage{hhline}
\usepackage[flushleft]{threeparttable}
\usepackage{array}
\usepackage{url}
\usepackage{latexsym}
\usepackage{float}
\usepackage{graphicx}
\floatstyle{plaintop}
\restylefloat{table}

\IEEEoverridecommandlockouts                              % This command is only needed if
\title{\LARGE \bf
A General Pipeline for 3D Detection of Vehicles
}

\author{Xinxin Du$^{1}$, Marcelo H. Ang Jr.$^{2}$, Sertac Karaman$^{3}$ and Daniela Rus$^{3}$% <-this % stops a space
\thanks{$^{1}$Xinxin Du is with the Singapore-MIT Alliance for Research and Technology (SMART) Centre, Singapore {\tt\small xinxin@smart.mit.edu}}%
\thanks{$^{2}$Marcelo H. Ang Jr. is with the National University of Singapore, Singapore {\tt\small mpeangh@nus.edu.sg}}%
\thanks{$^{3}$Sertac Karaman and Daniela Rus are with the Massachusetts Institute of Technology, Cambridge, MA, USA {\tt\small sertac@mit.edu rus@csail.mit.edu}}%
}
\begin{document}

\maketitle
%\thispagestyle{empty}
%\pagestyle{empty}

%%%%%%%%%%%%%%%%%%%%%%%%%%%%%%%%%%%%%%%%%%%%%%%%%%%%%%%%%%%%%%%%%%%%%%%%%%%%%%%%
\begin{abstract}
%2D vehicle detection has been well studied and developed. However, 3D information is needed for autonomous vehicle and 3D detection is not as well established as 2D.
Autonomous driving requires 3D perception of vehicles and other objects in the in environment. Much of the current methods support 2D vehicle detection. This paper proposes a flexible pipeline to adopt any 2D detection network and fuse it with a 3D point cloud to generate 3D information with minimum changes of the 2D detection networks. To identify the 3D box, an effective model fitting algorithm is developed based on generalised car models and score maps. A two-stage convolutional neural network (CNN) is proposed to refine the detected 3D box. This pipeline is tested on the KITTI dataset using two different 2D detection networks. The 3D detection results based on these two networks are similar, demonstrating the flexibility of the proposed pipeline. The results rank second among the 3D detection algorithms, indicating its competencies in 3D detection. 
\end{abstract}

%%%%%%%%%%%%%%%%%%%%%%%%%%%%%%%%%%%%%%%%%%%%%%%%%%%%%%%%%%%%%%%%%%%%%%%%%%%%%%%%
\section{INTRODUCTION}
\label{sec:Introduction}
Vision-based car detection has been well developed and widely implemented using deep learning technologies. The KITTI  \cite{Geiger2012CVPR} benchmark site reports that the state of the art algorithms are able to achieve $\sim 90\%$ average precision (AP). 

However, for autonomous vehicles, car detection in 2D images is not sufficient to provide enough information for the vehicle to perform planning and decision making due to the lack of depth data. For a robust and comprehensive perception system in autonomous vehicle, 3D car detection, including car dimensions, locations and orientations in the 3D world, is essential. However the state of the art for 3D car detection algorithms only achieve $62\%$ AP. Gaps still exist as compared to the 2D detection performance and the problem remains as challenging.

According to the types of input sources, the current algorithms for 3D vehicle detection can be categorised into four major groups, including (1) mono image based, (2) stereo image, (3) LiDAR (Light Detection and Ranging), and (4) fusion between mono image and Lidar.

Mono images lack the depth information to recover the 3D location of detected obstacles, therefore assumptions and approximations have to be made. Stereo image based approaches normally involve the construction of depth maps from stereo correspondence matching. The performance of this type of approach depends heavily on the depth map reconstruction and the accuracy drops as distance from the vehicle increases.

LiDAR, despite its high cost, is able to provide the most direct measurement of object location. But it lacks color information and it is always sparse which poses difficulties in classification. In order to make use of the full capabilities of LiDAR and camera, fusion approaches have been proposed in the literature. To make use of the deep CNN architecture, the point cloud needs to be transformed into other formats. In the process of transformation, information is lost.

The prior approaches for 3D vehicle detection are not as effective as those for 2D detection. Little attention has been paid to how to transfer the advantages and lessons learnt from 2D detection approaches to 3D detection approaches. Moreover, the field is lacking effective 3D detection approaches that enable the existing 2D approaches to provide 3D information. The state of the art 2D approaches can not be applied to autonomous vehicles which require 3D information.

\begin{figure*}[t]
\centering
	 \includegraphics[width=1.0\textwidth]{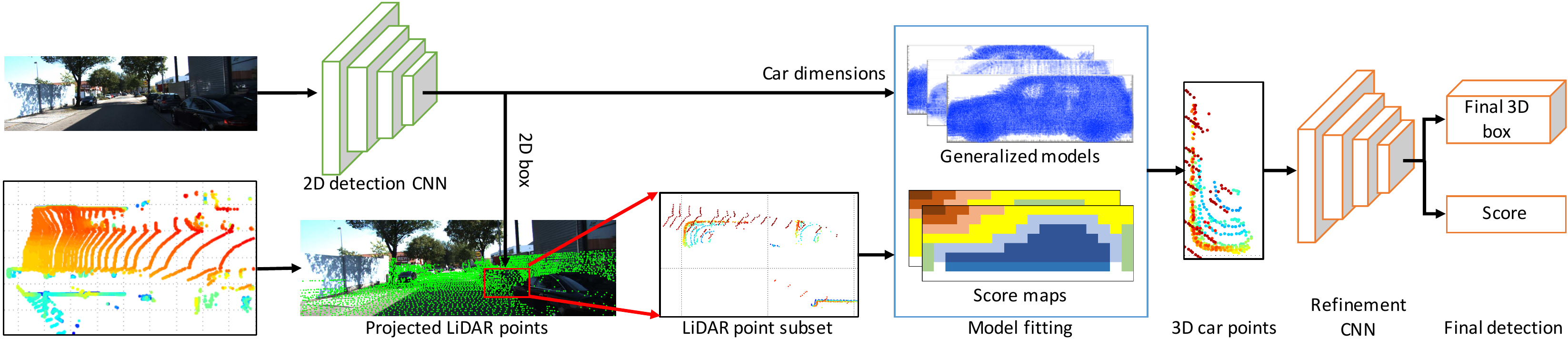}
\caption{General fusion pipeline. All of the point clouds shown are in 3D, but viewed from the top (bird's eye view). The height is encoded by color, with red being the ground. A subset of points is selected based on the 2D detection. Then, a model fitting algorithm based on the generalised car models and score maps is applied to find the car points in the subset and a two-stage refinement CNN is designed to fine tune the detected 3D box and re-assign an objectiveness score to it.}
\label{fig:General fusion pipeline}
\end{figure*}

In this paper, we propose a flexible 3D vehicle detection pipeline which can make use of any 2D detection network and provide accurate 3D detection results by fusing the 2D network with a 3D point cloud. The general framework structure is illustrated in Fig. \ref{fig:General fusion pipeline}. The raw image is passed to a 2D detection network which provides 2D boxes around the vehicles in the image plane. Subsequently, a set of 3D points which fall into the 2D bounding box after projection is selected.  With this set, a model fitting algorithm detects the 3D location and 3D bounding box of the vehicle. And then another CNN network, which takes the points that fit into the 3D bounding box as input, carries out the final 3D box regression and classification. It requires minimum efforts to modify the existing 2D networks to fit into the pipeline, adding just one additional regression term at the output layer to estimate the vehicle dimensions. The main contributions of the paper are:

\begin{enumerate}
\item A general pipeline that enables any 2D detection network to provide accurate 3D detection information.
\item Three generalised car models with score maps, which achieve a more efficient model fitting process.
\item A two-stage CNN that can further improve the detection accuracy.
\end{enumerate}

This pipeline has been tested using two outstanding 2D networks, PC-CNN \cite{du2017iros} and MS-CNN \cite{cai2016unified}. The 3D detection performances based on both networks were evaluated using the KITTI dataset \cite{Geiger2012CVPR}. We significantly lead the majority of the algorithms in both bird eye detection and 3D detection tasks. We also achieved comparable results to the current state of the art algorithm MV3D \cite{Chen2017CVPR} in both tasks.

\section{Related works}
\label{sec:related works}
This section reviews the works that are related to the proposed pipeline in details. It also highlights the differences between our proposal and the prior works.

\textbf{Mono Image Approaches:} In \cite{mousavian20163d}, a new network was designed to estimate the car dimensions, orientations and probabilities given a detected 2D box from an existing network. Using the criterion the perspective projection of a 3D box should fit tightly with the 2D box in the image, 3D box was recovered by using the estimated information. Similarly in DeepMANTA \cite{deepmanta_cvpr17}, the vehicle orientation and size were estimated from a deep CNN. Additionally, the network also estimated 36 locations of key points on the car in the image coordinates. A 2D/3D shape matching algorithm \cite{lepetit2009epnp} was applied to estimate vehicle 3D poses based on these 36 2D part locations. 

Another set of algorithms, e.g. \cite{xiang2015data}, \cite{xiang2017subcategory}, \cite{zeeshan2014cars} and \cite{zeeshan2013explicit}, defined 3D car models with occlusion patterns, carried out detection of the patterns in the 2D image and recovered the 3D occluded structure by reasoning through a MAP (maximum a posteriori) framework.

These approaches are sensitive to the assumptions made and the parameter estimation accuracy. As shown in the result section, our method outperforms them significantly.

\textbf{Stereo Image Approaches:} The depth map from stereo correspondence is normally appended to the RGB image as the fourth channel. The RGB-D image is passed to one or more CNNs in order to carry out the detection. In \cite{Pham2017}, Pham $et~al.$ proposed a two-stream CNN where the depth channel and RGB channel went through two separate CNN branches and were fused before the fully connected layers. 

\textbf{Lidar Approaches:} The common framework involves three steps: pre-processing (e.g. voxelization), segmentation and classification. A detailed review of LiDAR approaches can be found in \cite{pendleton2017perception}. Wang $et~al.$ \cite{wang2015voting} proposed a different approach where the point cloud was converted into 3D feature grids and a 3D detection window was slid through the feature grids to identify vehicles. In \cite{li2016vehicle}, the point cloud was converted into a 2D point map and a CNN was designed to identify the vehicle bounding boxes in the 2D point map. In \cite{li20163d}, the authors extended the approach of \cite{li2016vehicle} and applied 3D deep CNN directly on the point cloud. However, this approach is very time consuming and memory intensive due to the 3D convolutions involved. To improve, \cite{engelcke2017vote3deep} proposed a voting mechanism able to perform sparse 3D convolution.

\textbf{Fusion Approaches:} In \cite{eitel2015multimodal}, the sparse point cloud is converted to a dense depth image, which is similar to a stereo one. The RGB-D image was passed through a CNN for detection. In \cite{schlosser2016fusing}, the point cloud was converted into a three-channel map HHA which contains horizontal disparity, height above ground and angle in each channel. The resulting six-channel image RGB-HHA was processed by a CNN for detection of vehicles. However these two methods will not be able to output the 3D information directly from the network.

In oder to address this, MV3D (multi-view 3D) detection network proposed by Chen $et~al.$ \cite{Chen2017CVPR} included one more type of input generated from the point cloud, the bird's eye view feature input. This input has no projective loss as compared to the depth map, thus 3D proposal boxes can be generated directly. This approach has achieved the current state of the art in 3D vehicle detection. It generates 2D boxes from 3D boxes while ours generate 3D boxes from 2D boxes. And MV3D explores the entire point cloud while ours only focus on a few subsets of the point cloud, which is more efficient and saves computation power.

\textbf{2D Detection:} The proposed pipeline is flexible in regards to the choice of 2D detection networks. Only a slight change is required on the last fully connected layer of the network so that it is able to estimate the dimensions of the cars. Both \cite{mousavian20163d} and \cite{deepmanta_cvpr17} proposed ways to encode the car dimensions to the network. For better accuracy, the 2D detection networks proposed in \cite{Ren17CVPR}, \cite{deepmanta_cvpr17} and \cite{yang2016exploit} can be incorporated since they are the leading networks for 2D detection. For faster computation, the approaches presented in \cite{redmon2017yolo9000} and \cite{liu2016ssd} can be implemented. In this paper, we implement PC-CNN \cite{du2017iros} and MS-CNN \cite{cai2016unified} to demonstrate the flexibility of the pipeline.

\textbf{Model Fitting:} In \cite{xiang2015data}, Xiang $et~al.$ proposed 3D voxel patterns (3DVPs) as the 3D car model. 3DVPs encode the occlusion, self-occlusion and truncation information. A boosting detector was designed to identify the 3DVPs in the image, while \cite{xiang2017subcategory} implemented a sub-category awareness CNN for 3DVP detection. 

Deformable part-based models (DPM) can be found in \cite{pepik2012teaching}, \cite{forsyth2014object}, \cite{yebes2014supervised} and \cite{pepik2015multi}. Different classifiers were trained to detect DPM. Fidler $et~al.$ extended the DPM to a 3D cuboid model in \cite{fidler20123d} in order to allow reasoning in 3D. In \cite{zeeshan2013explicit}, \cite{zia2013detailed}, \cite{zeeshan2014cars} and \cite{deepmanta_cvpr17}, 3D wireframe models were used. Similarly, each wire vertex is encoded with its visibility.  

Due to the various vehicle types, sizes, and occlusion patterns, these prior approaches require a substantial number of models in order to cover all possible cases. In our approach, only three models are used and the occlusion pattern is assigned online when doing model fitting.

%\textbf{Fusion Pipeline:} The proposed fusion pipeline is different from the MV3D \cite{Chen2017CVPR}. In MV3D, the entire point cloud was converted to bird eye view input feature and 3D bounding box proposals were generated by convoluting the bird eye view input through CNN. The front view feature input and the RGB image go through another two separate CNNs. ROI pooling is performed on the three CNNs based on the 3D box proposals and its corresponding 2D boxes in the image. The three feature vectors given by ROI pooling operations are then fused to make 3D box regression and classification. The pipeline generates 2D boxes from 3D boxes while ours generate 3D boxes from 2D boxes. And MV3D explores the entire point cloud while ours only focus on a few subsets of the point cloud, which is more efficient and saves computation power.

%\textbf{2D to 3D:} In \cite{mousavian20163d}, the authors introduced a way to convert the 2D detection results to 3D information. The approach is different from ours. It uses mono image instead of fusing the image with point cloud. A new network was designed to estimate the car dimensions, orientations and probabilities given a detected 2D box. Using the criterion the perspective projection of a 3D box should fit tightly with the 2D box in the image, 3D box was recovered by using the estimated information. But it is sensitive to the accuracy of the 2D box and the estimated parameters. As shown in the result section, our method outperforms it significantly.

\section{Technical Approach}
%This section explains the model fitting and CNN 3D box regression in details. Subsets of the point cloud will be generated first based on the 2D boxes. From the transformation matrix between LiDAR and camera which can be obtained from calibration, the point cloud can be projected into the image plane. With the 2D bounding box, it is straightforward to get which of the projected points fall into the 2D box, subsequently the corresponding 3D points can be obtained.  
The input is an image. The first step is to generate 2D bounding boxes for the candidate vehicles. Secondly, these bounding boxes are used to select subsets of the point clouds, using the transformation between the camera and LiDAR. Due to the perspective nature of the camera, the 3D point subset may spread across a much larger area than the vehicle itself as shown in Fig.\ref{fig:General fusion pipeline}. This subset also contains a substantial number of non-vehicle points and points on neighbouring vehicles. All these artefacts add challenges to the 3D box detection.

\subsection{Car dimension estimation}
One additional regression layer is needed at the end of the given 2D detection network. This regression method was inspired by \cite{deepmanta_cvpr17} and \cite{mousavian20163d}. First the average dimensions for all the cars and vans in KITTI dataset is obtained. Let $[\bar h, ~\bar l, ~\bar w]$ denote height, length and width of the vehicle. The ground truth regression vector $\Delta_i^* = (\delta_h,~\delta_l,~\delta_w)$ is defined as:
\begin{align}
\delta_h = log(h^*/\bar h)~~~\delta_l = log(l^*/\bar l)~~~\delta_w = log(w^*/\bar w)
\label{eq:deltas}
\end{align}

The dimension regression loss is shown as:
\begin{align}
L_d(i) = \lambda_d C_i R(\Delta_i - \Delta_i^*)
\end{align}
where $\lambda_d$ is the weighting factor to balance the losses defined in the original network, e.g. classification loss, 2D box regression loss; $C_i$ is $1$ if the 2D box is a car and $0$ otherwise; $R$ is the smooth $L_1$ loss function defined in \cite{girshick2015fast} and $\Delta_i$ is the regression vector from the network.

To train the modified network, we can reuse the pre-trained weights from the original network for initialisation. Only a small part of the network needs to be re-trained while the rest can be kept as fixed during training. For example, in MS-CNN, we only re-trained the convolution layer and the fully connected layers after ROI pooling in the detection sub-network and in PC-CNN, we re-trained the GoogleNet layer, convolution layer and the fully connected layers after the De-convolution layer in the detection sub-network.

\subsection{Vehicle model fitting}
We first generate a set of 3D box proposals. For each proposal, the points within the 3D box are compared to the three generalised car models. The proposal with the highest score is selected for the two-stage CNN refinement.

The 3D box proposals are generated following the principle of RANSAC algorithm (random sample consensus). In each iteration, one point is selected randomly. A second point is randomly selected from points within the cube centred at the first point and with the side length of $1.5l$, where $l$ is the car length from the 2D CNN dimension estimation and $1.5$ compensates the estimation error. A vertical plane is derived from these two points. Any points with a distance to the plane less than a threshold are considered as inliers to the plane. A maximum 20 points are then randomly selected from the inliers. At each point, a second vertical plane, passing through that point and perpendicular to the first vertical plane, is derived. 

Along the intersection line between these two vertical planes, eight 3D boxes can be generated based on the estimated car width and length. Since the first vertical plane is visible, based on the view direction, four boxes are eliminated. At each of the remaining box locations, a new range is defined by expanding the box by 1.5 times along both $w$ and $l$ directions. The lowest point within the new range can be found and it determines the ground of the 3D box while the roof of the 3D box is set based on the height estimation. In summary, at each iteration, maximum 80 3D box proposals can be generated.

\begin{figure}[h]
\centering
    \subfloat[SUV]{
    \includegraphics[width=0.155\textwidth]{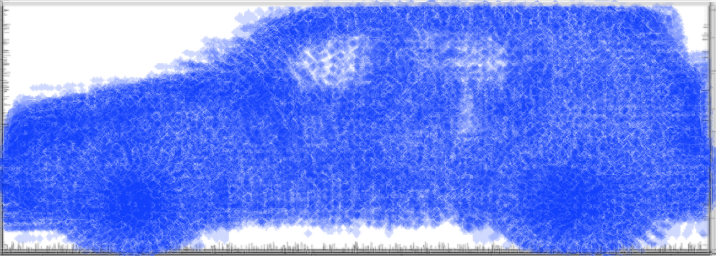}}
    \subfloat[Sedan]{
    \includegraphics[width=0.155\textwidth]{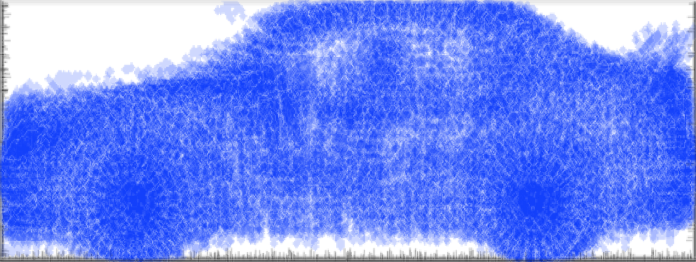}}
    \subfloat[Van]{
    \includegraphics[width=0.155\textwidth]{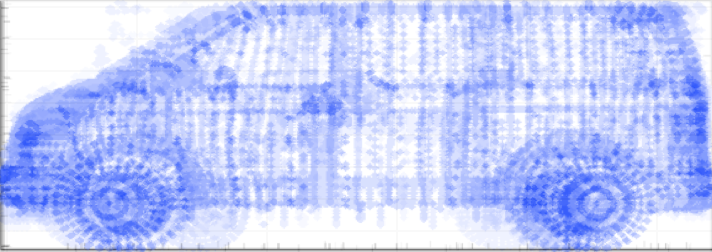}}
    \caption{Generalised car models}
    \label{fig:Generalised car models}
\end{figure}

Only three generalised car models are used for model fitting. They represent three categories of cars, SUVs, Sedans and Vans. Hatchback cars are considered to be SUVs. We observe that the relative distances/positions between different parts of a car do not vary significantly for different cars from the same category even with different sizes. This invariance indicates that if the cars under the same category are normalised to the same dimension $[h,~l,~w]$, their shapes and contours will be similar. We verified this and generalised the car models by normalising the cars in the 3D CAD dataset used in \cite{deepmanta_cvpr17}, \cite{fidler20123d} and \cite{chen2014beat}. Figure \ref{fig:Generalised car models} illustrates the side view of the point cloud plots for the three categories. Each plot is an aggregation of the points that are generated from the 3D CAD models, aligned to the same direction and normalised to the same dimension. The SUV/hatchback plot consists of points from 58 CAD models, the sedan plot consists of 65 point sets, and the van plot consists of points from 10 models.

Each aggregation is then voxelized to a $8\times 18 \times 10$ matrix along the $[h~l~w]$ direction. Each element in the matrix is assigned different scores based on its position. The elements representing the car shell/surface are assigned a score of $1$, indicating that 3D points in the model fitting process fall on the car surface will be counted towards the overall score. The elements inside or outside the car shell are assigned negative scores, and the further away they get from the car shell (either inwards or outwards), the smaller the assigned values. This indicates that no points should be detected from outside or inside the car by LiDAR and the overall score will be penalised for such detections. The elements at the bottom layer of the matrix are assigned a score of $0$. Points detected at the bottom layer could be the ground or car's tires, which are difficult to distinguish from each other. They will not be penalised or counted.

\begin{figure}[h!]
\centering
     \includegraphics[width=0.45\textwidth]{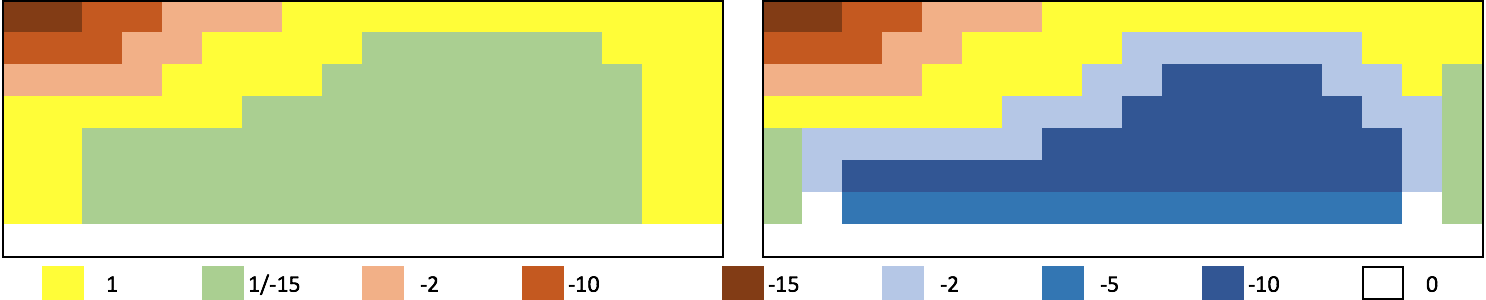}
    \caption{Score map (scores are indicated at bottom.)}
    \label{fig:Score map}
\end{figure}

Self-occlusion can be easily determined from the view direction. This is encoded online when doing the model fitting since view direction changes for different 3D box proposals. Negative scores are assigned to the car surface elements if they are self-occluded. Furthermore, for simplicity, only the four vertical facets are considered for self-occlusion analysis while car roof and bottom are not considered.

Two slices of the score assignment from the SUV category are shown in Fig. \ref{fig:Score map}, with the left image depicting the side facet and the right image illustrating the center slice. The car exterior and interior are indicated by orange and blue while the bottom is indicated white. Yellow and green refer to the shell/surface of the car, while green further indicates that those areas might be self-occluded.

Points within the 3D box proposals will be voxelised into $8 \times 18 \times 10$ grids and compared to the three potential vehicle models. Due to the orientation ambiguity, the grids are rotated around their vertical center axis by 180 degree and are then compared to the three models. Out of all the bounding box proposals, the one with the highest score is selected for the next step.

\subsection{Two-stage refinement CNN}
To further align the detected 3D box to the point cloud, we designed a two-stage CNN. In the literature, 3D CNNs are commonly used to process 3D point clouds, e.g. \cite{li2016vehicle}. However, these CNNs are extremely slow and memory intensive. In this paper, we found that 2D CNNs are sufficient.

With the points in a given 3D box, the first CNN outputs a new 3D box. A new set of points can be found within the new 3D box. The second CNN outputs a probability score based on the new set of points to indicate how likely these points represent an actual car.

However, point sets cannot be input to the CNN directly. We apply normalization and voxelization strategies in order to formalize the points in matrix form in order to fit to the CNN. Furthermore, consistent with 2D image detection cases \cite{bell2016inside}, \cite{cai2016unified}, a bounding box context is able to provide additional information to improve the detection accuracy. We also include the context of the 3D bounding box as input to the CNN.

Given a 3D box from the model fitting process, our pipeline expands it along its $h,~l,~w$ directions by $1.5, ~1.5,$ and $1.6$ times respectively to include its context. The points inside this expanded box are normalised and voxelised into a $24 \times 54 \times 32$ matrix. The matrix is sparse with $\sim 0.6\%$ occupied elements on average. As compared to the generalised model, we doubled the resolution of the voxelisation in order to reserve more spatial details and patterns of the distribution of the points. Note that the normalisation is anisometric, it has different scale ratios along different directions.

The backbones of the CNNs in both stages are based on the VGG-net with configuration D as described in \cite{simonyan2014very}. After each convolution, an ELU (exponential linear units) layer \cite{clevert2015fast}, instead of Re(ctified)LU layer, is adopted for a more stable training process. The first stage CNN has two parallel outputs, one for 3D box regression and the other for classification, while the second stage CNN only has one output, classification.
\begin{align}
\nonumber
\delta_{x_c}^* &= (X_c^*-X_c)/L~~~\delta_{y_c}^* = (Y_c^*-Y_c)/H~~~\delta_{z_c}^* = (Z_c^*-Z_c)/W \\ \nonumber
\delta_{x_l}^* &= (X_l^*-X_l)/L~~~\delta_{y_l}^* = (Y_l^*-Y_l)/H~~~~\delta_{z_l}^* = (Z_l^*-Z_l)/W \\ 
\delta_{w}^* &= log(W^*/W) 
\label{eq:deltas_3d}
\end{align}

The classification loss for both CNNs is $SoftMax$ loss and the 3D box regression loss is $SmoothL_1$ loss. The ground truth regression vector $\Delta_{3d}^*$ defined in (\ref{eq:deltas_3d}) has seven elements,  three for the center of the box, three for the left bottom corner and one for the width. It is just sufficient to recover the 3D bounding box from these seven elements. Due to the anisometric normalisation, a quartic polynomial needs to be solved. Note that across all the inputs, $X_{c/l}, ~Y_{c/l},~Z_{c/l},~L,~H,~W$ are all constant as all the 3D boxes are aligned and normalised to the same size.

\begin{table*}[t]
\centering
\caption{Average Precision benchmark for bird's eye view and 3D box based on KITTI validation set.}
\resizebox{0.95\textwidth}{!}{
\begin{tabular}{|c|c|c|c|c|c|c||c|c|c|c|c|c|}
    \hline
    \multirow{3}{*}{Algorithm} 	& \multicolumn{6}{c||}{Bird's Eye View} 	& \multicolumn{6}{c|}{3D Box} \\ \hhline{~------------}
    						& \multicolumn{3}{c|}{IoU = 0.5} 		& \multicolumn{3}{c||}{IoU = 0.7}  & \multicolumn{3}{c|}{IoU = 0.5} & \multicolumn{3}{c|}{IoU = 0.7}  \\ \hhline{~------------}
						& Easy & Moderate & Hard & Easy & Moderate & Hard & Easy & Moderate & Hard & Easy & Moderate & Hard \\ \hline
   $^*$Mono3D \cite{chen2016monocular}	& 30.50 & 22.39 & 19.16 &   5.22 & 5.19 & 4.13 & 25.19 & 18.20 & 15.52 & 2.53 & 2.31 & 2.31 \\ 
   $^*$3DOP \cite{xiang2015data}			& 55.04 & 41.25 & 34.55 & 12.63 & 9.49 & 7.59 & 46.04 & 34.63 & 30.09 & 6.55 & 5.07 & 4.10 \\
   $^*$$^*$Deep3DBox \cite{mousavian20163d}& 29.96 & 24.91 & 19.46 & 9.01 & 7.94 & 6.57 & 24.76 & 21.95 & 16.87 & 5.40 & 5.66 & 3.97 \\
   $^*$VeloFCN \cite{li2016vehicle}			& 79.68 & 63.82 & 62.80 & 40.14 & 32.08 & 30.47 & 67.92 & 57.57 & 52.56 & 15.20 & 13.66 & 15.98 \\
   $^*$MV3D \cite{Chen2017CVPR}			& 96.34 & 89.39 & 88.67 & 86.55 & 78.10 & 76.67 & 96.02 & 89.05 & 88.38 & 71.29 & 62.68 & 56.56 \\
   Ours (MS-CNN \cite{cai2016unified})	& 90.36 & 88.46 & 84.75 & 82.17 & 77.15 & 74.42 & 87.16 & 87.38 & 79.40 & 55.82 & 55.26 & 51.89 \\
   Ours (PC-CNN \cite{du2017iros})		& 88.31 & 83.74 & 79.62 & 83.61 & 77.36 & 69.61 & 87.69 & 79.92 & 78.65 & 57.63 & 51.74 & 51.39 \\
    \hline
    \multicolumn{13}{l}{\small $^*$ sources from \cite{Chen2017CVPR}} \\
    \multicolumn{13}{p{1.0\textwidth}}{\small $^*$$^*$ sources from \cite{mousavian20163d}, which uses different validation set, so its APs are calculated from the $1848$ common images with our validation set.}
\end{tabular}
}
\label{table:AP}
\end{table*}

Classification has two classes,  car and background. A 3D box is classified as positive when the IOU (intersection of union) between its bird's eye view box and the ground truth bird's eye view box is greater than a specific threshold. This threshold is 0.5 for the first stage CNN and 0.7 for the second. 0.7 is consistent with the criteria set by KITTI benchmark. The reason to set a lower threshold for the first stage is to train the network so that it is able to refine the boxes with IoU between 0.5 to 0.7 to a better position where the IoU may be greater than 0.7; otherwise the network will take those boxes as negative and will not be trained to refine them.

The training of the two networks is carried out independently as they do not share layers. The training batch size is 128, with $50\%$ being positive. Both CNNs are trained for 10K iterations with a constant learning rate of 0.0005.

\section{experiment results and discussion}
\label{sec:experiment results and discussion}
To verify the flexibility of our approach, the pipeline is tested using PC-CNN \cite{du2017iros} and MS-CNN \cite{cai2016unified}. The performance based on both networks is evaluated using the challenging KITTI dataset \cite{Geiger2012CVPR}, which contains 7481 images for training/validation and 7518 images for testing. The training/validation set has annotated ground truth for 2D bounding box in the image plane and 3D bounding box in real world. Following \cite{chen20153d}, we split the training/validation set into training and validation sub-sets. The training sub-set is purely used to train the car dimension regression and two-stage CNN while the validation sub-set is for evaluation only. KITTI divides the cars into easy, moderate and hard groups based on their visibilities. We follow this same convention for our evaluation. To further verify the performance of the proposed pipeline, we also tested it using our own autonomous vehicles.

$\textbf{Metrics:}$ The primary focus of this paper is on 3D detection, we do not evaluate the performance of the pipeline for 2D detection tasks. Following the evaluation metrics proposed in \cite{Chen2017CVPR}, we evaluate our proposal based on the Average Precession (AP) for bird's eye view boxes and for 3D boxes. The bird's eye view boxes are generated by projecting the 3D boxes on the same ground plane. The AP is calculated based on the IoU between the output boxes and the ground truth boxes, while in \cite{xiang2015data} and \cite{deepmanta_cvpr17}, the distance between two boxes are used. We feel that IoU is a more comprehensive index than distance, as it implicitly accounts for not only distance but also alignment and size.

\begin{figure*}[t]
\centering
     \includegraphics[width=0.32\textwidth]{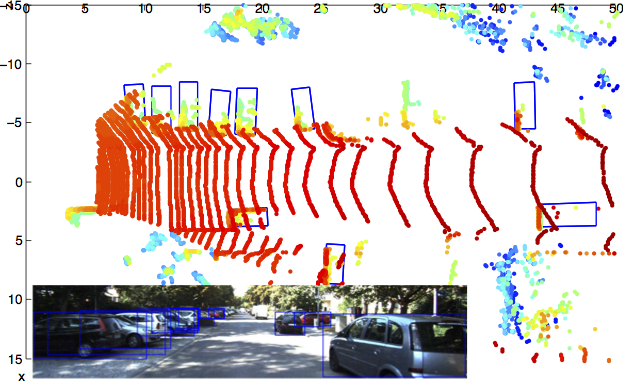} 
     \includegraphics[width=0.32\textwidth]{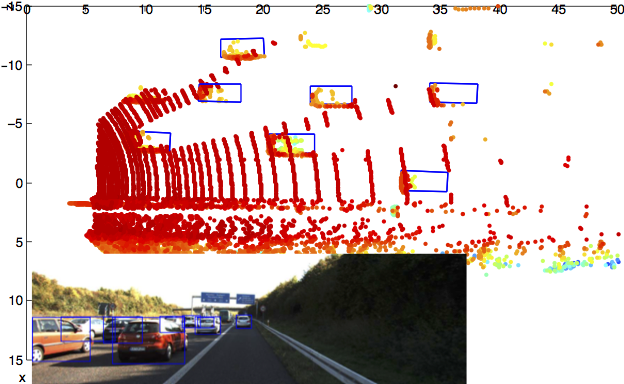}
     \includegraphics[width=0.32\textwidth]{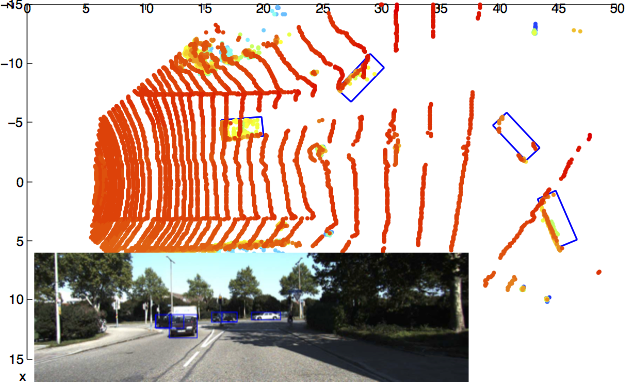}  \\
     \includegraphics[width=0.32\textwidth]{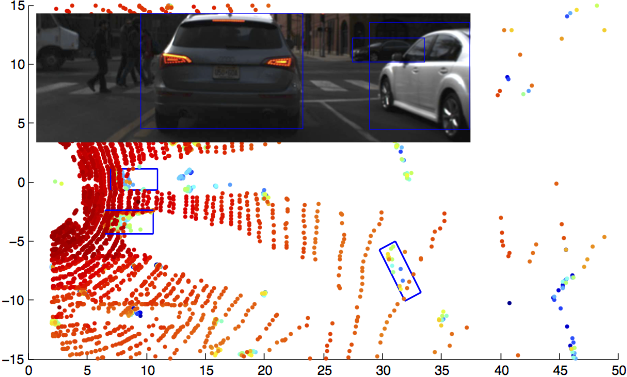} 
     \includegraphics[width=0.32\textwidth]{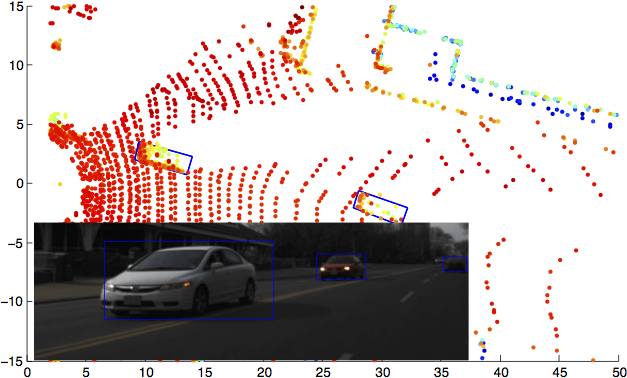} 
     \includegraphics[width=0.32\textwidth]{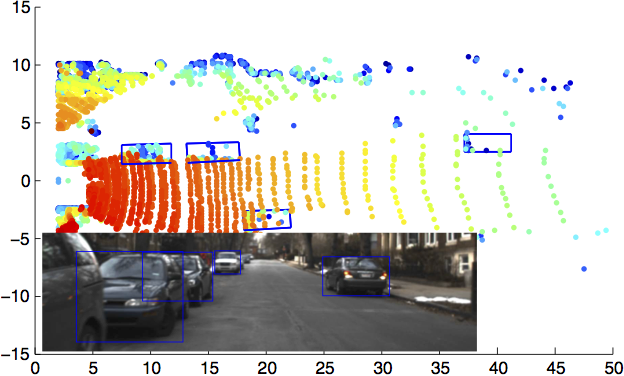} 
    \caption{Qualitative result illustration on KITTI data (top row) and Boston data (bottom row). Blue boxes are the 3D detection results.}
    \label{fig:qualitative results}
\end{figure*}

$\textbf{Bird's Eye View \& 3D Box AP:}$ We compare the outputs from our pipeline with other algorithms which can output 3D box information, including Mono3D \cite{chen2016monocular}, 3DOP \cite{xiang2015data} and Deep3DBox \cite{mousavian20163d} which use image data only, VeloFCN \cite{li2016vehicle} which uses LiDAR data only and MV3D \cite{Chen2017CVPR} which uses fusion.

The IoU threshold for true positive detection is set at 0.5 and 0.7. The left part of TABLE \ref{table:AP} shows the results from bird's eye view. In general, the point cloud based approaches all significantly lead the image-based approaches for both IoUs. Within the point cloud based approaches, our pipeline outperforms VeloFCN significantly but underperforms MV3D marginally. When $IoU = 0.5$, our performance with PC-CNN is about $7\%$ worse on average than MV3D and $5\%$ worse for MS-CNN. When $IoU=0.7$, the performances with PC-CNN and MS-CNN are both very close to MV3D except for the performance with PC-CNN for the hard group ($7\%$ worse than MV3D).

The 3D box detection comparisons are listed in the right part of TABLE \ref{table:AP}. Similarly for both IoU thresholds, our method significantly outperforms all the approaches with the single exception of MV3D. On average, the overall performance is about $10\%$ worse than MV3D for both $IoU=0.5$ and $0.7$ except that the performance with MS-CNN for moderate group at $IoU=0.5$ is only $1.6\%$ less than MV3D.

We only use point clouds to generate the 3D box and do not take any color information from the image into account. Comparing it to VeloFCN, which also only takes point clouds as inputs, shows the effectiveness of our approach, processing the point cloud as subsets instead of as a whole. Comparing to MV3D, image color information is necessary to further boost the performance of our pipeline. One possible solution is to extract the feature layer which is right before the $ROI~pooling$ layer in the 2D detection CNN. Based on the 3D box from the model fitting process, we could find its corresponding 2D box in the image plane and carry out $ROI~pooling$ on the extracted feature layer in order to extract the feature vector. Then fuse the feature vector with the one from the refinement CNN to output the final 3D box and its probability.

$\textbf{Flexibility Anlysis:}$ The comparison between the two approaches using the proposed pipeline verifies the flexibility of our pipeline. PC-CNN and MS-CNN have different network structures. But both approaches achieve comparable AP for the two tasks and IoU thresholds. Furthermore, the two-stage refinement CNN was trained based on the pipeline with PC-CNN and re-used in the pipeline with MS-CNN without any further tuning on the network. This further confirms the flexibility and adaptability of our proposed pipeline.

$\textbf{Car Dimension Regression Impact:}$  We show the impact from the car dimension regression on the original 2D detection CNN in TABLE \ref{table:Impact on 2D CNN}. Similarly, AP is populated for the 2D detection task in image plane. Following KITTI, the IoU threshold is set at $0.7$. The left part shows the performance of the original 2D detection CNN while the right part indicates the results after appending the car dimension regression term. The impact is not very significant for both networks, and it even improves the performance marginally for some groups.

\begin{table}[h]
\centering
\caption{Impact on the original 2D detection CNN from appending the car dimension regression term.}
\resizebox{0.48\textwidth}{!}{
\begin{tabular}{|c|c|c|c||c|c|c|}
\hline
 \multirow{2}{*}{2D Detection} 	& \multicolumn{3}{c||}{Original} 	& \multicolumn{3}{c|}{With Dimension Regression} \\ \hhline{~------}
 					& Easy & Moderate & Hard & Easy & Moderate & Hard \\ \hline
MS-CNN				& 91.64 & 89.95 & 79.55 & 93.98 & 89.92 & 79.69 \\ 
PC-CNN				& 94.62 & 89.60 & 79.97 & 90.22 & 89.03 & 81.64 \\		
\hline
\end{tabular}
}
\label{table:Impact on 2D CNN}
\end{table}

$\textbf{Ablation Study:}$ To analyse the effectiveness of the steps involved in the 3D box generation, the AP is calculated after each step (model fitting, first stage CNN and second stage CNN) for both bird's eye view and 3D box tasks. For this study, the IoU threshold is set to $0.5$. Since the results based on MS-CNN and PC-CNN are quite comparable, only PC-CNN results are presented in TABLE \ref{table:Ablation study}. 

The results from the model fitting are not as good as the final oens, but they are better than all the image based algorithms  in TABLE \ref{table:AP} and comparable to VeloFCN. This indicates that the model fitting algorithm can work properly. 

\begin{table}[h]
\centering
\caption{Ablation study based on KITTI validation set. Numbers indicate AP with IoU threshold at 0.5.}
\resizebox{0.48\textwidth}{!}{
\begin{tabular}{|c|c|c|c||c|c|c|}
\hline
 \multirow{2}{*}{Step} 	& \multicolumn{3}{c||}{Bird's Eye View} 	& \multicolumn{3}{c|}{3D Box} \\ \hhline{~------}
 					& Easy & Moderate & Hard & Easy & Moderate & Hard \\ \hline
Model fitting			& 77.71 & 73.27 & 70.06 & 56.32 & 51.33 & 47.40  \\ 
First CNN				& 88.16 & 83.60 & 79.65 & 87.51 & 79.76 & 78.81 \\		
Second CNN			& 88.31 & 83.74 & 79.62 & 87.69 & 79.92 & 78.65 \\ \hline
\end{tabular}
}
\label{table:Ablation study}
\end{table}

With the first CNN, the detection performance is improved significantly in both bird's eye view and 3D box tasks. The improvement is $\sim 10\%$ and $\sim 30\%$ respectively. This shows that although only 2D convolution is used and the input 3D matrix is very sparse, the network is still very powerful and effective to locate the 3D box. The improvement from the second CNN is insignificant since it is not designed to regress the 3D box. It is designed to reshuffle the probability of the 3D box from the first CNN.

%Model fitting (iou0.5, bird) 77.71 73.27 70.06; (iou0.5 3d) 56.32 51.33 47.40
%Model fitting (iou0.7, bird) 42.83 36.59 34.38; (iou0.7 3d) 05.92 05.26 04.54
%1st CNN       (iou0.5, bird) 88.16 83.60 79.65; (iou0.7 3d) 87.51 79.76 78.81
%1st CNN       (iou0.7, bird) 82.68 76.80 69.28; (iou0.7 3d) 56.67 51.10 51.28
%2nd CNN      (iou0.5, bird) 88.31 83.74 79.62; (iou0.5 3d) 87.69 79.92 78.65; 
%2nd CNN      (iou0.7, bird) 83.61 77.36 69.61; (iou0.7 3d) 57.63 51.74 51.39;

% original MS-CNN 2D detection 		91.64 89.95 79.55
% MS-CNN 2D detection with dim reg 	93.98 89.92 79.69

% original PC-CNN 2D detection		94.62 89.60 79.97
% PC-CNN 2D detection with dim reg 	90.22 89.03 81.64

$\textbf{Qualitative Results:}$ The first row in Fig. \ref{fig:qualitative results} shows some of the 3D detection results by applying our pipeline with PC-CNN on KITTI validation dataset. We also tested it using our own dataset collected at Boston USA. The setup of the data collection vehicle is similar to KITTI,  with differences in the relative positions between the LiDAR, camera and the car. We applied the pipeline, which is trained based on KITTI training dataset, directly on the Boston data without any fine-tuning of the network weights. The system still works as shown in the second row of Fig. \ref{fig:qualitative results}. It shows the generalisation capability of the proposed pipeline and indicates its potentials in executing 3D vehicle detection in real situations beyond a pre-designed dataset. Interested readers may refer to the link for video illustrations (\url{https://www.dropbox.com/s/5hzjvw911xa5mye/kitti_3d.avi?dl=0}).

\section{CONCLUSIONS}
\label{sec:conclusions}

In this paper we propose a flexible 3D vehicle detection pipeline which is able to adopt the advantages of any 2D detection networks in order to provide 3D information. The effort to adapt the 2D networks to the pipeline is minimal. One additional regression term is needed at the network output to estimate vehicle dimensions. The pipeline also takes advantage of point clouds in 3D measurements. An effective model fitting algorithm based on generalised car models and score maps is proposed to fit the 3D bounding boxes from the point cloud. Finally a two-stage CNN is developed to fine tune the 3D box. The outstanding results based on two different 2D networks indicate the flexibility of the pipeline and its capability in 3D vehicle detection.

\section*{ACKNOWLEDGMENT}
This research was supported by the National Research Foundation, Prime Minister's Office, Singapore, under its CREATE programme, Singapore-MIT Alliance for Research and Technology (SMART) Future Urban Mobility (FM) IRG. We are grateful for their support.

\bibliographystyle{ieeetr}	
\bibliography{ICRA_2018}

\end{document}